\documentclass[lettersize,journal]{IEEEtran}
\usepackage{amsmath,amsfonts}
\usepackage{algorithmic}
\usepackage{algorithm}
\usepackage{array}
\usepackage[caption=false,font=normalsize,labelfont=sf,textfont=sf]{subfig}
\usepackage{textcomp}
\usepackage{stfloats}
\usepackage{url}
\usepackage{verbatim}
\usepackage{graphicx}
\usepackage{cite}
\usepackage{booktabs}
\usepackage{algorithm}
\usepackage{algorithmic}
\usepackage{xcolor}
\usepackage[cmyk]{xcolor}
\usepackage{caption}  

\definecolor{cmykRoad}{cmyk}{0,1,0,0}               
\definecolor{cmykSidewalk}{cmyk}{0,0.25,0.2,0}       
\definecolor{cmykParking}{cmyk}{0,0,0,0.17}          
\definecolor{cmykOtherGround}{cmyk}{0,0.35,1.0,0}    
\definecolor{cmykBuilding}{cmyk}{0,0.75,0.75,0.35}   
\definecolor{cmykCar}{cmyk}{1,0,0,0}                 
\definecolor{cmykTruck}{cmyk}{1,0,0,0.5}             
\definecolor{cmykBicycle}{cmyk}{1,1,0,0}             
\definecolor{cmykMotorcycle}{cmyk}{0,1,0,0.5}        
\definecolor{cmykOtherVeh}{cmyk}{0,0,0,0.34}         
\definecolor{cmykVegetation}{cmyk}{1,0,1,0.5}        
\definecolor{cmykTrunk}{cmyk}{0,0,1,0.5}             
\definecolor{cmykTerrain}{cmyk}{1,0,1,0}             
\definecolor{cmykPerson}{cmyk}{0,1,0,0}              
\definecolor{cmykBicyclist}{cmyk}{0,1,1,0}           
\definecolor{cmykMotorcyclist}{cmyk}{0,0,0,0.5}      
\definecolor{cmykFence}{cmyk}{0.25,0.06,0,0.1}       
\definecolor{cmykPole}{cmyk}{0,0,1,0}                
\definecolor{cmykTrafficSign}{cmyk}{0,0.35,1.0,0}    
\definecolor{lightblue}{RGB}{173, 216, 230}
\usepackage{amssymb}

\usepackage{algorithm}
\usepackage{algorithmic}
\usepackage{amsmath}
\usepackage{multirow}

\begin{document}

\title{CurriFlow: Curriculum-Guided Depth Fusion with Optical Flow-Based Temporal Alignment for 3D Semantic Scene Completion}

\author{Jinzhou Lin,\thanks{Jinzhou Lin and Jie Zhou contribute equally.} 
Jie Zhou, 
Wenhao Xu,
Rongtao Xu,
 Changwei Wang, Shunpeng Chen,Kexue Fu, Yihua Shao, \\Li Guo, Shibiao Xu
$^{\dag}$,~\IEEEmembership{Member,~IEEE,} \thanks{Shibiao Xu is the corresponding author (shibiaoxu@bupt.edu.cn).}\thanks{Jinzhou Lin, Jie Zhou, Wenhao Xu, Shunpeng Chen, Li Guo and Shibiao Xu are with the School of Artificial Intelligence, Beijing University of Posts and Telecommunications, China.}
\thanks{Rongtao Xu is with the Spatialtemporal AI}
\thanks{Yihua Shao is with the Institute of Automation, Chinese Academy of Sciences, China.}
\thanks{Changwei Wang and Kexue Fu are with the Key Laboratory of Computing Power Network and Information Security, Ministry of Education, Shandong Computer Science Center (National Supercomputer Center in Jinan), Qilu University of Technology (Shandong Academy of Sciences), also with Shandong Provincial Key Laboratory of Computing Power Internet and Service Computing, Shandong Fundamental Research Center for Computer Science, Jinan, 250014, China}}

\markboth{IEEE Transactions on Circuits and Systems for Video Technology}%
{Jinzhou Lin and Jie Zhou \MakeLowercase{\textit{et al.}}: CurriFlow: Curriculum-Guided Depth Fusion with Optical Flow-Based Temporal Alignment for 3D Semantic Scene Completion}

\maketitle

\begin{abstract}
Semantic Scene Completion (SSC) aims to infer complete 3D geometry and semantics from monocular images, serving as a crucial capability for camera-based perception in autonomous driving. However, existing SSC methods relying on temporal stacking or depth projection often lack explicit motion reasoning and struggle with occlusions and noisy depth supervision.
We propose CurriFlow, a novel semantic occupancy prediction framework that integrates optical flow-based temporal alignment with curriculum-guided depth fusion. CurriFlow employs a multi-level fusion strategy to align segmentation, visual, and depth features across frames using pre-trained optical flow, thereby improving temporal consistency and dynamic object understanding. To enhance geometric robustness, a curriculum learning mechanism progressively transitions from sparse yet accurate LiDAR depth to dense but noisy stereo depth during training, ensuring stable optimization and seamless adaptation to real-world deployment. Furthermore, semantic priors from the Segment Anything Model (SAM) provide category-agnostic supervision, strengthening voxel-level semantic learning and spatial consistency.
Experiments on the SemanticKITTI benchmark demonstrate that CurriFlow achieves state-of-the-art performance with a mean IoU of 16.9, validating the effectiveness of our motion-guided and curriculum-aware design for camera-based 3D semantic scene completion.
\end{abstract}

\begin{IEEEkeywords}
3D Semantic Occupancy Prediction, Autonomous Driving, Temporal Alignment, Curriculum Learning.
\end{IEEEkeywords}

\section{Introduction}
\IEEEPARstart{S}{SC} aims to infer complete 3D geometry and semantic information from partial observations such as monocular images, serving as a key task in visual perception for autonomous driving and robotics~\cite{tian2023occ3d,li2023voxformer,cao2022monoscene,han2025multimodal}. 
Traditional SSC methods primarily rely on depth estimation or voxel projection to reconstruct 3D scenes~\cite{yu2024context}. 
Although depth maps enhance spatial awareness, they suffer from significant limitations under occlusion and discontinuities: missing depth leads to incomplete voxels, while inaccurate depth can cause geometric distortion and semantic confusion, thereby degrading structural consistency and scene integrity.

In recent years, several studies have attempted to alleviate these problems by introducing temporal information. 
Historical frames often contain geometric and texture details missing in the current frame, which can help compensate for voxel incompleteness caused by depth estimation~\cite{wang2024not,li2024bevformer}. 
However, existing approaches often adopt simple frame stacking or pose-based projection, lacking explicit modeling of object-level motion. 
As a result, they struggle to accurately capture dynamic scene changes and tend to introduce temporal blurring and semantic misalignment during feature fusion. 
\textbf{This raises a critical question: Can object-level motion be leveraged to integrate depth geometry with temporal information, thereby enhancing spatial understanding?}
\begin{figure}
    \centering
    \includegraphics[width=\linewidth]{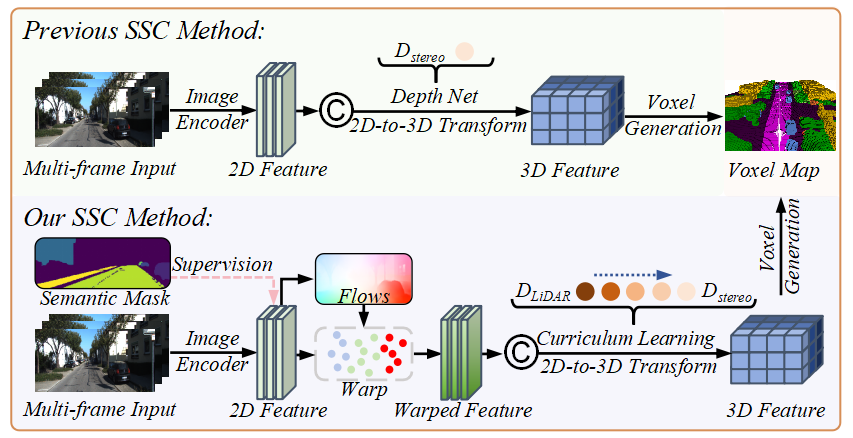}
    \caption{Comparison between the previous and our SSC methods. (Top) The previous approach simply stacks multi-frame inputs and projects 2D features into voxel space through depth-based 2D-to-3D transformation, which suffers from depth noise and temporal misalignment. (Bottom) Our method explicitly maps and aligns temporal features via optical flow, guided by semantic supervision and curriculum-based depth fusion, enabling temporally consistent and semantically complete voxel generation.}
    \label{fig:first}
\end{figure}
To address this issue, we propose to explicitly align cross-frame features using optical flow. 
Optical flow captures pixel-level motion displacement and provides rich temporal cues for modeling dynamic objects~\cite{xu2022gmflow,cho2024flowtrack}. 
By leveraging optical flow-guided cross-frame feature alignment, our method effectively reduces error propagation in occluded regions and enhances the continuity and stability of 3D reconstruction. 
Considering that optical flow estimation may become unreliable under illumination variation or texture degradation, CurriFlow introduces a confidence-guided occlusion masking mechanism that adaptively weights low-confidence regions, thereby improving the robustness of temporal alignment.

Meanwhile, depth fusion remains crucial for achieving accurate geometric reconstruction. 
While stereo-predicted depth is dense but noisy, LiDAR depth is sparse yet highly precise. 
To balance these complementary properties, CurriFlow adopts a curriculum-guided depth fusion strategy, where training starts with sparse but accurate LiDAR supervision and gradually transitions to dense stereo supervision. 
This curriculum design enables a smooth shift from stable optimization to pure camera-based inference, improving both learning stability and real-world adaptability.

Furthermore, with the advancement of foundation vision models, large-scale pre-trained models now offer strong semantic priors for downstream tasks~\cite{chen2025sage}. 
We incorporate the Segment Anything Model (SAM), whose category-agnostic segmentation masks provide high-level semantic guidance, ensuring globally consistent supervision for voxel-level semantic learning and improving spatial coherence.

In summary, \textbf{CurriFlow} achieves more robust and temporally consistent 3D semantic scene completion by integrating \textbf{optical flow-guided temporal feature fusion}, \textbf{curriculum-guided depth learning}, and \textbf{semantic prior distillation}. Our main contributions are summarized as follows:
\begin{itemize}
    \item[(1)] We propose \textbf{CurriFlow}, a unified framework for camera-based SSC that combines optical flow-guided temporal alignment with curriculum-guided depth fusion, effectively addressing occlusion and motion-induced misalignment.
    
    \item[(2)] We design a \textbf{confidence-aware and optical flow-guided temporal feature fusion mechanism} that explicitly models motion cues and adaptively aligns features across frames, improving temporal consistency and robustness under dynamic conditions.
    
    \item[(3)] We propose a \textbf{curriculum-guided depth learning scheme} that progressively shifts from sparse but accurate LiDAR supervision to dense stereo depth during training, while relying solely on camera input during inference. 
    This design ensures geometric stability throughout training and guarantees full camera-only compatibility at test time.
    
    \item[(4)] Extensive experiments on the \textbf{SemanticKITTI} and \textbf{SSCBench-KITTI360} demonstrate that CurriFlow achieves state-of-the-art performance, validating the effectiveness of the proposed motion-guided and curriculum-aware design.
\end{itemize}

\section{Related Work}
\subsection{Semantic Scene Completion}

SSC aims to predict volumetric semantic occupancy for both observed and occluded areas in a 3D scene. In autonomous driving, lightweight and accurate SSC methods are essential for real-time deployment.

Early work such as SSCNet~\cite{song2017semantic} used RGB-D input and 3D CNNs for indoor scene completion, but its dense voxel-based architecture is unsuitable for large-scale outdoor scenes. With the release of outdoor SSC datasets like SemanticKITTI~\cite{behley2019semantickitti} and nuScenes~\cite{caesar2020nuscenes}, research has gradually shifted toward sparse LiDAR and monocular RGB-based solutions. Vision-only methods have gained traction due to their low hardware cost and easier deployment. MonoScene~\cite{cao2022monoscene} first demonstrated that RGB-only input could be used for SSC via a 3D U-Net architecture. TPVFormer~\cite{huang2023tri} introduced tri-perspective view fusion with attention-based lifting, enhancing Bird’s-Eye View (BEV) reasoning. VoxFormer~\cite{li2023voxformer} further leverages sparse BEV queries and a cross-modality transformer to improve performance under occlusion and long-range scenarios. Meanwhile, multi-modal fusion methods have also seen rapid progress. OccDepth~\cite{miao2023occdepth} uses LiDAR-based ground-truth depth as supervision to guide pseudo-depth generation. CGFormer~\cite{yu2024context} incorporates conditional mechanisms and BEV-guided fusion for robust reasoning in dynamic scenes. Other works such as Occ3D~\cite{tian2023occ3d}explore cross-modal fusion, hierarchical feature aggregation, and 3D-aware representations to improve SSC performance.

However, due to the complexity and variability of real-world scenes, relying solely on a single-frame image for scene reconstruction is far from sufficient. Incorporating an optical flow module to temporally align multi-frame images before reconstruction is undoubtedly a more effective solution.
\begin{figure*} 
    \centering
    \includegraphics[width=\textwidth]{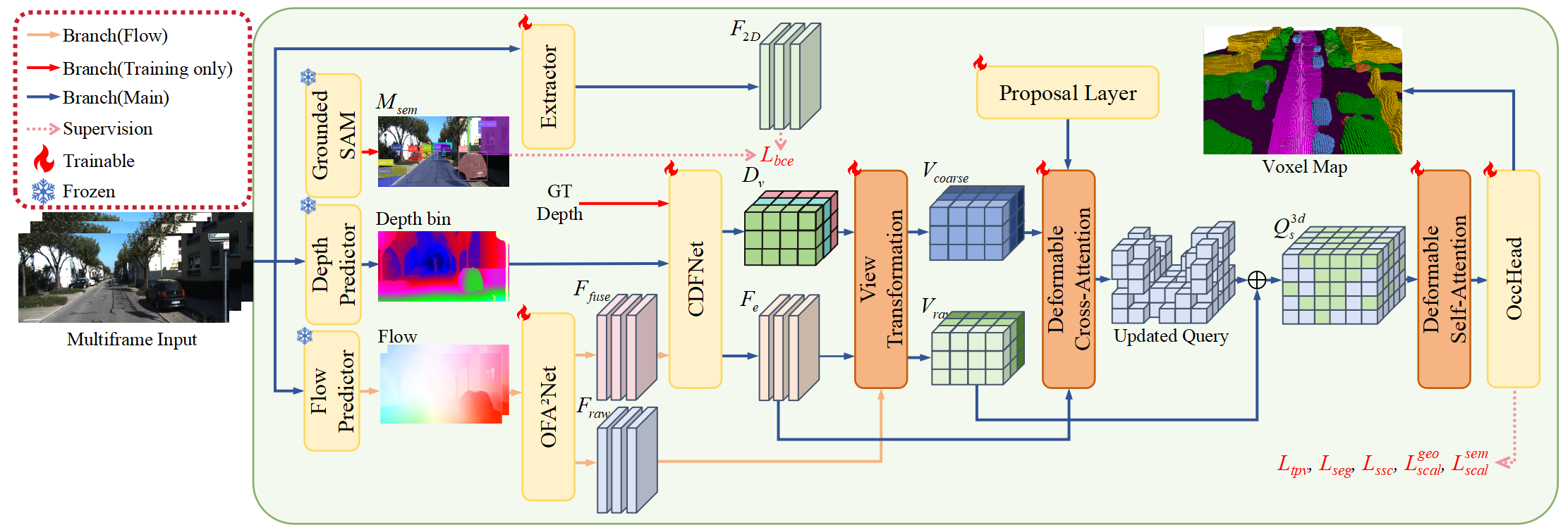}
    \caption{The overall CurriFlow architecture inputs three frames for depth, optical flow, and Grounded-SAM segmentation, extracting depth, flow, and instance masks. Instance masks assist the semantic loss to improve segmentation. The extracted image features, depth, and flow are temporally aligned by OFA\textsuperscript{2}Net, fused via CDFNet, and further encoded by a spatial encoder for voxel-context modeling and multi-scale semantic aggregation.}
    \label{fig:pipeline}
\end{figure*}
\subsection{Optical Flow}

Optical flow estimation is a fundamental task in computer vision, aiming to estimate pixel-wise motion between consecutive frames. Traditional methods such as Horn–Schunck~\cite{bruhn2005lucas} and Lucas–Kanade~\cite{baker2004lucas} provide accurate estimates under small displacements, but degrade under large motion or occlusion.

Deep learning has significantly advanced optical flow performance. FlowNet~\cite{dosovitskiy2015flownet} introduced the first end-to-end CNN model for optical flow, later improved by FlowNet2~\cite{ilg2017flownet} with multi-scale stacking. PWC-Net~\cite{sun2018pwc} became a popular choice due to its pyramid warping and cost volume design. RAFT~\cite{teed2020raft} leveraged iterative refinement and all-pairs correlation for high accuracy on challenging datasets.More recently, Transformer-based methods brought global receptive fields and improved matching. GMA~\cite{jiang2021learning} uses attention to model long-range dependencies, while GMFlow~\cite{xu2022gmflow} treats flow as global correspondence matching. FlowFormer~\cite{huang2022flowformer} unifies cost volume construction and update using Transformer blocks, improving robustness under occlusion and large displacement. Beyond motion estimation, optical flow has proven beneficial in multi-frame vision tasks. FlowTrack~\cite{cho2024flowtrack} improves multi-frame object tracking using flow-guided aggregation. FGFA~\cite{zhu2017flow} enhances video object detection with flow-based feature warping. In semantic segmentation.

Despite progress in temporal modeling, optical flow remains underexplored in SSC. Explicit motion reasoning via flow priors could enhance temporal consistency and improve reconstruction in dynamic, occluded scenes.

\section{Methodology}
\subsection{Overview}
Recent progress in camera-based SSC highlights the importance of temporal consistency, geometric reliability, and spatial coherence. However, most prior works address these aspects in isolation, treating temporal alignment, depth completion, and voxel refinement as separate components. 

We propose a unified framework, \textbf{CurriFlow}, that integrates these factors under the principle of \textit{temporal–geometric consistency}. Specifically, optical flow-based temporal alignment ensures motion coherence across frames, curriculum-guided depth fusion stabilizes geometric estimation during training, and deformable voxel refinement enhances 3D spatial completeness and semantic consistency. The overall pipeline is illustrated in Figure~\ref{fig:pipeline}.

\subsection{OFA\textsuperscript{2}Net}

\begin{figure}[h]
    \centering
    \includegraphics[width=\linewidth]{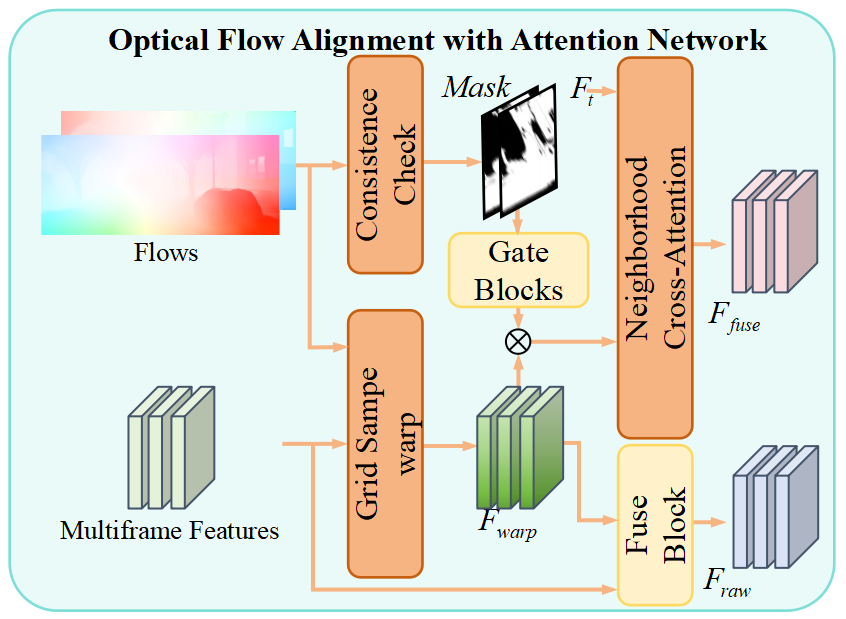}
    \caption{The OFA\textsuperscript{2}Net aligns historical frame features via attention-based grid sampling, producing initial features $F_{\text{raw}}$. Occlusion mask gates error suppression, and neighborhood cross-attention fuses features into $F_{\text{fuse}}$.}
    \label{fig:flowblock}
\end{figure}

Inspired by optical flow, we propose \textbf{OFA\textsuperscript{2}Net}, a temporal alignment framework that aligns historical features to the current frame with sub-pixel accuracy, enabling coherent fusion across frames, shown in Figure \ref{fig:flowblock}

 Given the current frame image $I_t$ and historical frames $\{I_{t-i}\}_{i=1}^{n}$, we first employ a pretrained optical flow estimation network to estimate the bidirectional optical flows between each pair of frames, denoted as $\{Flow_{fwd},Flow_{bwd}\}_{i=1}^{n}$. We then warp the image features of the historical frame $F_{t-i}$ to the current frame using the backward flow $Flow_{bwd_i}$, formulated as:
 \begin{equation}
     F_{warp}^{t-i \rightarrow t} = Warp(F_{t-i},\ Flow_{bwd_i})
 \end{equation}
where $warp(\cdot)$ denotes the feature warping operation implemented via $grid\_sample$, where a sampling grid is constructed based on the optical flow.

In autonomous driving, the relative motion between the camera and the environment causes objects to shift out of view, creating occluded regions with unmatched pixels during warping.
To identify such regions, we apply a forward-backward consistency check, which assumes that the forward and backward optical flows should be opposite in direction and equal in magnitude. The computation procedure is illustrated in algorithm \ref{alg:fwd_bwd_check}.
\begin{algorithm}[ht]
\caption{Forward-Backward Consistency Check}
\label{alg:fwd_bwd_check}
\begin{algorithmic}[1]
\REQUIRE Forward flow $\mathbf{F}_{fwd}$, backward flow $\mathbf{F}_{bwd}$, constants $\alpha, \beta$
\ENSURE Occlusion masks $\mathbf{M}_{fwd}, \mathbf{M}_{bwd}$

\STATE $\mathbf{mag} \leftarrow \|\mathbf{F}_{fwd}\|_2 + \|\mathbf{F}_{bwd}\|_2$
\STATE $\hat{\mathbf{F}}_{bwd} \leftarrow \mathrm{warp}(\mathbf{F}_{bwd}, \mathbf{F}_{fwd})$
\STATE $\hat{\mathbf{F}}_{fwd} \leftarrow \mathrm{warp}(\mathbf{F}_{fwd}, \mathbf{F}_{bwd})$
\STATE $\mathbf{T} \leftarrow \alpha \cdot \mathbf{mag} + \beta$
\STATE $\mathbf{M}_{fwd} \leftarrow (\|\mathbf{F}_{fwd} + \hat{\mathbf{F}}_{bwd}\|_2 > \mathbf{T})$
\STATE $\mathbf{M}_{bwd} \leftarrow (\|\mathbf{F}_{bwd} + \hat{\mathbf{F}}_{fwd}\|_2 > \mathbf{T})$
\RETURN $\mathbf{M}_{fwd}, \mathbf{M}_{bwd}$
\end{algorithmic}
\end{algorithm}

The occlusion mask is used as a confidence weight in the Mask Gate module to filter warped historical features, retaining only high-confidence regions and reducing unreliable temporal information in the current frame.

Next, we enhance the current frame representation by applying the Neighborhood Cross-Attention (NCA) mechanism~\cite{hassani2023neighborhood}, where the current frame features serve as the query, and the filtered historical features act as the key and value:
\begin{equation}
    F_{fuse} = NCA(F_t,\ F_{warp}^{mask})
\end{equation}

Meanwhile, the unfiltered warped features $F_{warp}$ are concatenated with the current features $F_t$ to form a residual input $F_{raw}$ for subsequent processing. \textbf{The temporally coherent features $F_{fuse}$ provide stable motion-aware cues that guide the subsequent depth fusion process, ensuring that geometric estimation is aligned with temporal dynamics.}

\subsection{CDFNet}

While temporal alignment stabilizes motion consistency, geometric accuracy remains sensitive to depth noise. To address this, we propose \textbf{CDFNet}, a curriculum-guided depth fusion module that progressively transitions from LiDAR supervision to stereo-based estimation during training. As raw LiDAR depth is too sparse for dense image alignment, we first apply a depth completion network ~\cite{zhang2023completionformer} to transform sparse LiDAR points into dense depth maps. Meanwhile, in line with prior studies, stereo depth is predicted using a pre-trained stereo matching network MobileStereoNet~\cite{Shamsafar_2022_WACV}. These two sources are then used to generate the fused training-time depth input.

We define the fused depth during training as:
\begin{equation}
D_{\text{fused}} = \lambda(t)\cdot D_{dense} + \left(1 - \lambda(t)\right)\cdot D_{stereo},
\end{equation}
where the completed ground-truth depth is obtained by:
\begin{equation}
D_{\text{dense}} = \mathcal{DC}(D_{gt}),
\end{equation}
where $\mathcal{DC}(\cdot)$ denotes a depth completion model that reconstructs dense depth maps from sparse LiDAR measurements and \( \lambda(t) \) is a decaying weight function that gradually shifts from LiDAR-based supervision to stereo estimation as training progresses.

\begin{figure}[ht]
\centering
\includegraphics[width=\linewidth]{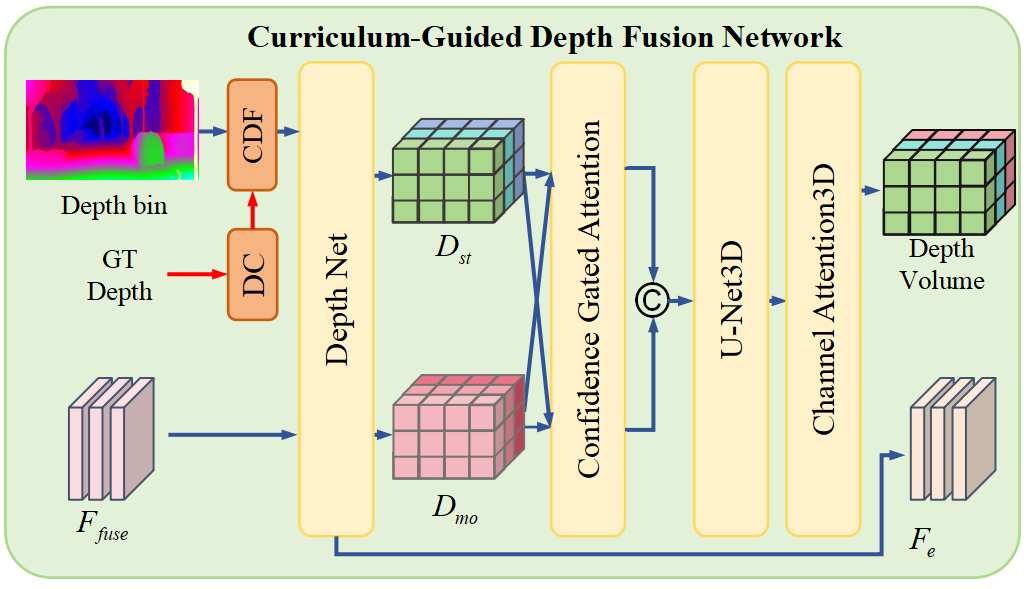}
\caption{The CDFNet fuses stereo and monocular volumes through bi-directional attention, to produce depth volumes and $F_e$. Notice that DC and CDF is only active during training. DC: depth completion, CDF: curriculum-guided fusion }
\label{fig:cdf}
\end{figure}

The completed dense depth is concatenated with fused image features $F_{fuse}$ and processed by a depth feature network to generate monodepth $D_{mo}$ and stereodepth $D_{st}$ volumes. To enable robust cross-modal fusion, we apply symmetric confidence attention modules, where each volume is refined under the guidance of the other, enhancing complementary cues and mitigating depth uncertainty.
\begin{equation}
    Q_{st},K_{mo},V_{mo} = Conv3D_{q,k,v}(D_{st},D_{mo},D_{mo})
\end{equation}

\begin{equation}
A_{st \rightarrow mo} = Softmax \left( \frac{Q_{st} \cdot K_{mo}^\top}{\sqrt{d}} \right)
\end{equation}
\begin{equation}
\hat{V}_{mo} = A_{st \rightarrow mo} \cdot \mathbf{V}_{mo},
P_{conf} = Softmax(\tilde{V}_{st})
\end{equation}
\begin{equation}
V_{mo}^{weighted} = P_{conf} \odot \hat{V}_{mo}, 
V_{st}^{weighted} = P_{conf} \odot \hat{V}_{st}.
\end{equation}
The above formulas are defined as $CGAttention3D(\cdot)$.
A symmetric operation is applied to obtain $V_{st}^{weighted}$ via:
\begin{equation}
    \hat{V}_{st} = CGAttention3D(D_{mo}, D_{st}).
\end{equation}

The $V_{st}^{weighted}$ and the $V_{mo}^{weighted}$ are concatenated ,then fed into a 3D convolution layer for initial feature fusion. This is followed by the U-Net, composed of multiple 3D convolutional layers, and skip connections. U-Net captures multi-scale contextual information, enhancing both local detail and global structure awareness.

We further apply a channel attention module (CA3D)~\cite{hu2018squeeze} to model inter-channel dependencies and boost semantic discriminability. Finally, a convolutional head projects the fused features into the final depth volume $D_{v}$. textbf{These geometry-consistent depth features from CDFNet act as reliable priors for voxel-level reasoning in the subsequent stage, bridging 2D temporal perception and 3D structural learning.} The module is illustrated in Figure \ref{fig:cdf}.

\subsection{Voxel Generation}
To achieve spatially coherent 3D reconstruction, we follow a two-stage deformable attention paradigm inspired by VoxFormer~\cite{li2023voxformer}. In the first stage, we treat the initial coarse voxel representation $V_{coarse}$ generated by the Lift-Splat-Shoot (LSS)~\cite{philion2020lift} module as the query to attend to relevant image features:
\begin{equation}
V_{coarse}=LSS(D_v,F_e),
\end{equation}
\begin{equation}
V_{raw}=LSS(D_v,F_{raw})
\end{equation}

To enable spatially adaptive projection of 2D image features into the 3D voxel space, we employ a Deformable Cross-Attention (DCA) module~\cite{xia2022vision}. Guided by the proposal indices, DCA samples multi-scale image features around voxel locations, generating context-aware voxel embeddings:

\begin{equation}
    Q_s = DCA(Proposal,V_{coarse},F_e)
\end{equation}
Here, $Proposal$ denotes sparse sampling positions output from a lightweight proposal layer, $F_e$ is the projected feature, and $Q_s$ is the updated voxel query.

To improve temporal consistency, we integrate $V_{raw}$ from OFA\textsuperscript{2}Net, which encodes optical-flow-guided information, filling in missing structures due to viewpoint changes and enhancing the query representation.
\begin{equation}
    Q_{s}^{3d}=Q_s + v_{raw}
\end{equation}

In the second stage, we apply Deformable Self-Attention (DSA)~\cite{xia2022vision} entirely within the 3D voxel space:
\begin{equation}
V_{s}^{3d}=DSA(Q_s+V_{raw})
\end{equation}
DSA allows each voxel to attend to both local and distant neighbors, enhancing object integrity and spatial context, thereby improving the fine-grained completeness and semantic coherence of voxel representations. \textbf{The proposed CurriFlow unifies temporal alignment, geometric fusion, and spatial reasoning under a single principle of temporal–geometric consistency. Each component reinforces the others, forming a coherent framework that significantly improves robustness and interpretability in camera-based 3D semantic scene completion.}


\begin{table*}[!ht]
\centering
\renewcommand{\arraystretch}{1}
\resizebox{\textwidth}{!}{%
\begin{tabular}{|c|c|cccccccccccccccccccc|c|}
\hline
& &&\multicolumn{18}{c}{\textbf{Semantic Occupancy Prediction}} &&  \\ 
Method&Input&IoU&
\rotatebox{90}{\textcolor{cmykRoad}{\rule{0.3cm}{0.3cm}} road (15.3\%)} & 
\rotatebox{90}{\textcolor{cmykSidewalk}{\rule{0.3cm}{0.3cm}} sidewalk (11.1\%)} & 
\rotatebox{90}{\textcolor{cmykParking}{\rule{0.3cm}{0.3cm}} parking (1.1\%)} & 
\rotatebox{90}{\textcolor{cmykOtherGround}{\rule{0.3cm}{0.3cm}} other-ground (0.6\%)}& 
\rotatebox{90}{\textcolor{cmykBuilding}{\rule{0.3cm}{0.3cm}} building (14.4\%)} & 
\rotatebox{90}{\textcolor{cmykCar}{\rule{0.3cm}{0.3cm}} car (3.9\%)} & 
\rotatebox{90}{\textcolor{cmykTruck}{\rule{0.3cm}{0.3cm}} truck (0.3\%)} & 
\rotatebox{90}{\textcolor{cmykBicycle}{\rule{0.3cm}{0.3cm}} bicycle (0.1\%)} & 
\rotatebox{90}{\textcolor{cmykMotorcycle}{\rule{0.3cm}{0.3cm}} motorcycle (0.1\%)} & 
\rotatebox{90}{\textcolor{cmykOtherVeh}{\rule{0.3cm}{0.3cm}} other-veh. (0.2\%)} & 
\rotatebox{90}{\textcolor{cmykVegetation}{\rule{0.3cm}{0.3cm}} vegetation (39.3\%)} & 
\rotatebox{90}{\textcolor{cmykTrunk}{\rule{0.3cm}{0.3cm}} trunk (0.5\%)} & 
\rotatebox{90}{\textcolor{cmykTerrain}{\rule{0.3cm}{0.3cm}} terrain (9.2\%)} & 
\rotatebox{90}{\textcolor{cmykPerson}{\rule{0.3cm}{0.3cm}} person (0.1\%)} & 
\rotatebox{90}{\textcolor{cmykBicyclist}{\rule{0.3cm}{0.3cm}} bicyclist (0.1\%)} & 
\rotatebox{90}{\textcolor{cmykMotorcyclist}{\rule{0.3cm}{0.3cm}} motorcyclist (0.1\%)} & 
\rotatebox{90}{\textcolor{cmykFence}{\rule{0.3cm}{0.3cm}} fence (3.9\%)} & 
\rotatebox{90}{\textcolor{cmykPole}{\rule{0.3cm}{0.3cm}} pole (0.3\%)} & 
\rotatebox{90}{\textcolor{cmykTrafficSign}{\rule{0.3cm}{0.3cm}} traffic-sign (0.1\%)} &
 mIoU
\\ \hline

MonoScene  & S &34.2&54.7&27.1&24.8&5.7&14.4&18.8&3.3&0.5&0.7&4.4&14.9&2.4&19.5&1.0&1.4&0.4&11.1&3.3&2.1&11.1\\
TPVFormer & S &34.3&55.1&27.2&27.4&6.5&14.8&19.2&3.7&1.0&0.5&2.3&13.9&2.6&20.4&1.1&2.4&0.3&11.0&2.9&1.5&11.3\\
SurroundOcc&S&34.7&56.9&28.3&30.2&6.8&15.2&20.6&1.4&1.6&1.2&4.4&14.9&3.4&19.3&1.4&2.0&0.1&11.3&3.9&2.4&11.9\\
OccFormer  & S &34.5&55.9&30.3&31.5&6.5&15.7&21.6&1.2&1.5&1.7&3.2&16.8&3.9&21.3&2.2&1.1&0.2&11.9&3.8&3.7&12.3\\
IAMSSC&S&43.7&54.0&25.5&24.7&6.9&19.2&21.3&3.8&1.1&0.6&3.9&22.7&5.8&19.4&1.5&2.9&0.5&11.9&5.3&4.1&12.4\\
DepthSSC&S&44.6&55.6&27.3&25.7&5.8&20.5&21.9&3.7&1.4&1.0&4.2&23.4&7.6&21.6&1.3&2.8&0.3&12.9&5.9&6.2&13.1\\
VoxFormer-S&S&43.0&53.9&25.3&21.1&5.6&19.8&20.8&3.5&1.0&0.7&3.7&22.4&7.5&21.3&1.4&2.6&0.2&11.1&5.1&4.9&12.2\\
CGFormer &S&45.3&64.8&32.2&20.7&0.4&23.9&33.7&10.8&3.1&3.1&7.7&26.4&7.5&38.8&2.6&2.7&0.0&9.6&10.8&7.2&16.2\\ \hline
VoxFormer-T & T& 44.0&54.8&26.4&15.5&0.7&17.6&25.8 &5.6 &0.6&0.5&3.8&24.4&5.1&29.9&1.8&3.3&0.0&7.6 &7.1&4.2 &12.4\\
HASSC-T&  T &44.6&57.2&29.1&19.9&1.3&20.2&27.3&17.1&1.1&1.1&8.8&27.0&7.7&33.9&2.3&4.1&0.0&7.9&9.2&4.8&14.7\\
SGN&  T&45.4 & 59.0 & 30.1 & 19.4 & 0.2 & 23.9 & 32.5 & 9.7 & 0.4 & 0.1 & 5.2 & 28.3 & 8.6 & 34.9 & 0.8 & 0.2& 0.0 & 8.8 & 12.1 & 6.9 & 14.8
\\
H2GFormer-T& T&44.7&57.0&29.4&21.7&0.3&20.5&28.2&6.8&0.9&0.9&9.3&27.4&7.8&36.3&1.2&0.1&0.0&7.9&9.8&5.8&14.3\\ \hline
CurriFlow & T &\textbf{45.5}&\textbf{66.4}&\textbf{33.0}&\textbf{23.0}&0.1&\underline{21.3}&\textbf{33.8}&\textbf{18.3}&\textbf{3.1}&\textbf{3.9}&\textbf{11.7}&\underline{27.5}&\underline{7.9}&\textbf{39.7}&\textbf{2.8}&1.1&0.0&\textbf{10.7}&\underline{11.2}&\underline{6.7}&\textbf{16.9}\\
\hline
\end{tabular}%
}
\caption{Comparison on SemanticKITTI validation set. S: Single-frame input, T: Multi-frame input. \textbf{Bold} indicates the best performance among temporal methods, while \underline{underlined} marks the second-best.}
\label{table:comparison}
\end{table*}
\subsection{OccEncoder}
The OccEncoder follows the same architectural design as the CGFormer~\cite{yu2024context}, consisting of a local 3D encoder and a global TPV encoder. Specifically, the local branch adopts a 3D ResNet-based backbone to capture fine-grained spatial structures and geometric cues within the voxel space, enabling accurate modeling of local occupancy patterns. In parallel, the global branch utilizes a Swin Transformer-based TPV (Three-Plane View) encoder to aggregate long-range contextual information across orthogonal projections.
Both encoders operate in parallel, and their feature representations are adaptively fused through a learnable weighting mechanism, which balances the contributions of local geometry and global semantics. This parallel-weighted fusion effectively integrates complementary information and enhances the expressiveness of voxel representations.
\subsection{Semantic Distillation Module}

\begin{table*}[!h]

\centering
\renewcommand{\arraystretch}{1}
\resizebox{\textwidth}{!}{%
\begin{tabular}{|c|c|ccccccccccccccccccc|c|}
\hline
& &\multicolumn{18}{c}{\textbf{Semantic Occupancy Prediction}} &&  \\ 
Method&Input&IoU&

\rotatebox{90}{\textcolor{cyan}{\rule{0.3cm}{0.3cm}} car (2.85\%)} & 
\rotatebox{90}{\textcolor{blue}{\rule{0.3cm}{0.3cm}} bicycle (0.01\%)} & 
\rotatebox{90}{\textcolor{purple}{\rule{0.3cm}{0.3cm}} motorcycle (0.01\%)} & 
\rotatebox{90}{\textcolor{teal}{\rule{0.3cm}{0.3cm}} truck (0.16\%)} & 
\rotatebox{90}{\textcolor{darkgray}{\rule{0.3cm}{0.3cm}} other-vehicle (5.75\%)} & 
\rotatebox{90}{\textcolor{magenta}{\rule{0.3cm}{0.3cm}} road (14.98\%)} & 
\rotatebox{90}{\textcolor{magenta}{\rule{0.3cm}{0.3cm}} person (0.02\%)} &
\rotatebox{90}{\textcolor{lightgray}{\rule{0.3cm}{0.3cm}} parking (2.31\%)} & 
\rotatebox{90}{\textcolor{pink}{\rule{0.3cm}{0.3cm}} sidewalk (6.43\%)} & 
\rotatebox{90}{\textcolor{orange}{\rule{0.3cm}{0.3cm}} other-ground (2.05\%)} & 
\rotatebox{90}{\textcolor{brown}{\rule{0.3cm}{0.3cm}} building (15.67\%)} & 
\rotatebox{90}{\textcolor{lightblue}{\rule{0.3cm}{0.3cm}} fence (0.96\%)} & 
\rotatebox{90}{\textcolor{green}{\rule{0.3cm}{0.3cm}} vegetation (41.99\%)} & 
\rotatebox{90}{\textcolor{lime}{\rule{0.3cm}{0.3cm}} terrain (7.10\%)} & 
\rotatebox{90}{\textcolor{yellow}{\rule{0.3cm}{0.3cm}} pole (0.22\%)} & 
\rotatebox{90}{\textcolor{orange}{\rule{0.3cm}{0.3cm}} traffic-sign (0.06\%)} &
\rotatebox{90}{\textcolor{red}{\rule{0.3cm}{0.3cm}} other-struck. (4.33\%)} & 
\rotatebox{90}{\textcolor{gray}{\rule{0.3cm}{0.3cm}} other-obj. (0.28\%)} & 
 mIoU
\\ \hline
MonoScene  & S &37.9&  19.3 &0.4 &0.6&8.0 &2.0& 48.4 &0.9& 11.4 &28.1 &3. &32.9 &3.5 &26.2& 16.6 &6.9& 5.7 &4.2 &3.1 &12.3\\ 
TPVFormer &S &40.2&21.6 & 1.1   & 1.4 & 8.1 & 2.6 & 52.9& 2.4 & 11.9& 31.1 & 3.8 & 34.8 & 4.8& 30.1& 17.5 & 7.5 & 5.9 & 5.5& 2.7&13.6 \\ 
OccFormer  & S &40.2&  22.6& 0.7& 0.3& 9.9 &3.8& 54.3&2.8& 13.4& 31.5& 3.6& 36.4& 4.8 &31.0& 19.5& 7.8& 8.5 &6.9& 4.6& 13.8\\ 
 DepthSSC&S&40.9& 21.9& 2.4 &4.3 &11.5 &4.6 &50.9&2.9 & 12.9 &30.3& 2.5 &37.3& 5.2 &29.6 &21.6 &5.9&7.7& 5.2& 3.5&14.3 \\ \hline
VoxFormer-T & T&38.8&17.8&1.2&0.9& 4.6& 2.1&47.0&1.6& 9.7&27.2&2.9&31.2&4.9&28.9&14.7& 6.5& 6.9& 3.8 &2.4 &11.9\\ 
Symphonies&T&44.1&30.0&1.9&5.9&25.1&12.1&54.9&8,2&13.8&32.8&6.9&35.1&8.6&38.3&11.5&14.0& 9.6&14.4&11.3 &18.6\\
   \hline
CurriFlow & T &\textbf{47.5}&\underline{29.2}&\textbf{3.4} &\underline{4.4}&\underline{14.2}&\underline{7.2}&\textbf{63.8} &\underline{6.6}  &\textbf{17.5} &\textbf{40.6}   &\underline{5.1}&\textbf{41.6}&\textbf{8.7} &\underline{37.9}   &\textbf{23.7} &\textbf{15.6}  &\textbf{18.5} &\underline{9.9} &\underline{7.1}&\textbf{19.7}\\ \hline
\end{tabular}%
}
\caption{Comparison on SSCBench-KITTI360 test set. S: Single-frame input, T: Multi-frame input. \textbf{Bold} indicates the best performance among temporal methods, while \underline{underlined} marks the second-best.}
\label{table:comparison1}
\end{table*}

To facilitate semantic knowledge transfer under weak supervision, we introduce a Semantic Distillation Branch that injects object-level priors from pretrained foundation models to enhance voxel-level semantic reasoning, especially in RGB-only settings.

Specifically, we employ a lightweight 2D segmentation head that predicts a multi-channel semantic probability map aligned with the image resolution, based on features extracted by the image backbone. During training, this head is supervised using soft pseudo-labels generated by Grounded-SAM~\cite{ren2024grounded}, following a \textbf{prediction-level distillation} strategy. This allows the model to absorb rich semantic cues—such as object shapes and boundaries—without requiring full 3D annotations.

To further refine the network’s sensitivity to structural details, we introduce a Boundary-aware Distillation Loss consisting of:

\begin{itemize}
    \item \textbf{Category-level supervision:} standard cross-entropy loss for semantic alignment;
    \item \textbf{Shape-preserving supervision:} Dice loss to encourage region-level consistency;
    \item \textbf{Boundary-aware supervision:} combined Dice and binary cross-entropy loss on edge maps to improve boundary localization.
\end{itemize}

Importantly, this distillation branch is only active during training and can be removed during inference, ensuring no additional runtime cost.

\subsection{Training Loss}
To effectively supervise semantic scene completion in 3D voxel space, we design a multi-branch loss that jointly enforces geometric accuracy, semantic discrimination, cross-view consistency, and boundary precision. The overall training objective consists of three complementary components.

\subsubsection{Voxel-Level Supervision}
Following MonoScene~\cite{cao2022monoscene}, we employ multi-scale voxel supervision to enhance both geometric and semantic consistency. Specifically, 
\begin{itemize}
    \item \textbf{Geometric Scale Loss} ($L_{\text{scal}}^{\text{geo}}$) penalizes incorrect foreground/background occupancy predictions across hierarchical resolutions;
    \item \textbf{Semantic Scale Loss} ($L_{\text{scal}}^{\text{sem}}$) provides class-aware semantic guidance at multiple voxel scales;
    \item \textbf{Cross-Entropy Loss} ($L_{\text{ce}}$) is applied at full resolution to refine voxel-level semantic boundaries.
\end{itemize}
Together, these losses promote structure-aware semantic learning while maintaining spatial coherence across scales.

\subsubsection{Semantic Distillation Branch}
The semantic distillation module is optimized with a hybrid boundary-aware objective that aligns semantic priors from SAM with model predictions. It comprises:
\begin{itemize}
    \item \textbf{Cross-Entropy Loss}: Enforces pixel-level agreement between predicted logits and SAM-derived soft labels;
    \item \textbf{Dice Loss}: Maintains region-level consistency and alleviates class imbalance;
    \item \textbf{Boundary Loss}: Combines Dice and Binary Cross-Entropy terms on edge maps to enhance boundary localization.
\end{itemize}
This branch strengthens semantic sharpness and improves 2D–3D feature alignment.

\subsubsection{TPV-Based Cross-View Loss}
To encourage view-consistent learning, ground-truth voxels are projected onto three orthogonal planes and aligned with TPV features extracted from the Swin Transformer. A class-weighted cross-entropy loss enforces semantic consistency, assigning higher weights to distant voxels to improve long-range supervision.

\subsubsection{Overall Objective}
The total training loss integrates the three components as:
\begin{equation}
    \mathcal{L}_{total} = \lambda_1 \mathcal{L}_{voxel} + \lambda_2 \mathcal{L}_{distill} + \lambda_3 \mathcal{L}_{TPV},
\end{equation}
where $\lambda_1$, $\lambda_2$, and $\lambda_3$ balance the contributions of each term. This joint optimization ensures geometrically consistent, semantically precise, and view-aligned scene completion.

\begin{table}[t]
\centering
\scriptsize
\setlength{\tabcolsep}{3pt}
\renewcommand{\arraystretch}{0.9}

\resizebox{\linewidth}{!}{
\begin{tabular}{@{}cccc@{}}
\toprule
Type & Name & GPU Memory Consumption (MB) & Latency (s)  \\
\midrule
\multirow{3}{*}{Module} 
 & OFANet & 394.83 & --  \\
 & CDFNet & 3093.99 & --  \\
 & Distillation Branch & 15.62 & --  \\
\midrule
\multirow{2}{*}{Method} 
 & CGFormer & 19299& 1.82  \\
 & CurriFlow &21280  & 2.24  \\
\bottomrule
\end{tabular}}
\caption{GPU usage and efficiency comparison across modules and methods.}
\label{tab:gpu}
\end{table}
\section{Experiment}
\subsection{Setup}
We evaluate \textbf{CurriFlow} on the SemanticKITTI and SSCBench-KITTI360 datasets following the official data splits, and report all quantitative results on the test sets. Our framework is implemented in PyTorch and trained for 25 epochs on four RTX 3090 GPUs.

For fair comparison and stable training, we adopt standard pre-trained visual and depth backbones as feature extractors, while all temporal fusion, confidence estimation, and curriculum depth modules are designed and trained by us. Specifically, a high-resolution visual encoder is employed for image feature extraction, and a lightweight stereo depth network is used to generate dense depth inputs. Optical flow predictions are obtained from a pre-trained flow estimation module to provide temporal motion cues, which are further refined within our confidence-aware temporal fusion block. Sparse LiDAR supervision is densified through a depth completion network only during training to support the proposed curriculum-guided depth fusion strategy. 
All backbone parameters are frozen during training, and the CurriFlow components are optimized end-to-end. We report the GPU memory consumption and latency for each module and method in the Table \ref{tab:gpu}

 \subsection{Comparision with Other methods}

We present the results tested on SemanticKITTI in the Table \ref{table:comparison}. CurriFlow achieves a mIoU of 16.9 and an IoU of 45.4. For both mIoU and IoU, we outperform all other methods. From the perspective of individual categories, our method shows strong performance on moving objects (e.g., car: 28.2 → 33.8, truck: 6.8 → 18.3, other-vehicle: 9.3 → 11.7). Meanwhile, CurriFlow also achieves significant improvements on long-tail categories (e.g., person, motorcycle), further demonstrating the effectiveness of our approach.

It is worth noting that, regardless of whether optical flow alignment or other temporal modeling mechanisms are applied, almost all temporal methods consistently show low performance on the "other-ground" class in the SemanticKITTI dataset. This phenomenon is not caused by any specific model design but is closely related to the intrinsic nature of this class. Specifically, \textit{other-ground} typically appears in transition areas between roads and vegetation or in distant, sparsely observed regions, where geometric structures are weak and texture cues are insufficient, making it difficult to establish reliable temporal correspondences. In addition, its boundaries are ambiguously annotated, and the voxel proportion is extremely low, making it vulnerable to class imbalance during training. During temporal fusion, its features are often overwhelmed by dominant neighboring categories such as \textit{road} or \textit{vegetation}, further diminishing its discriminative power. Hence, this degradation reflects inherent limitations of the dataset and class definition rather than a flaw in temporal modeling. \textbf{Improving the IoU of representative low-texture and boundary-ambiguous classes, such as \textit{other-ground}, will be an important direction for future research.}
To demonstrate the generalization ability of the model, We conducted evaluations on SSCBench-KITTI360, as shown in the Table \ref{table:comparison1}.
\subsection{Qualitative Results}
\begin{figure*} 
    \centering
    \includegraphics[width=\textwidth]{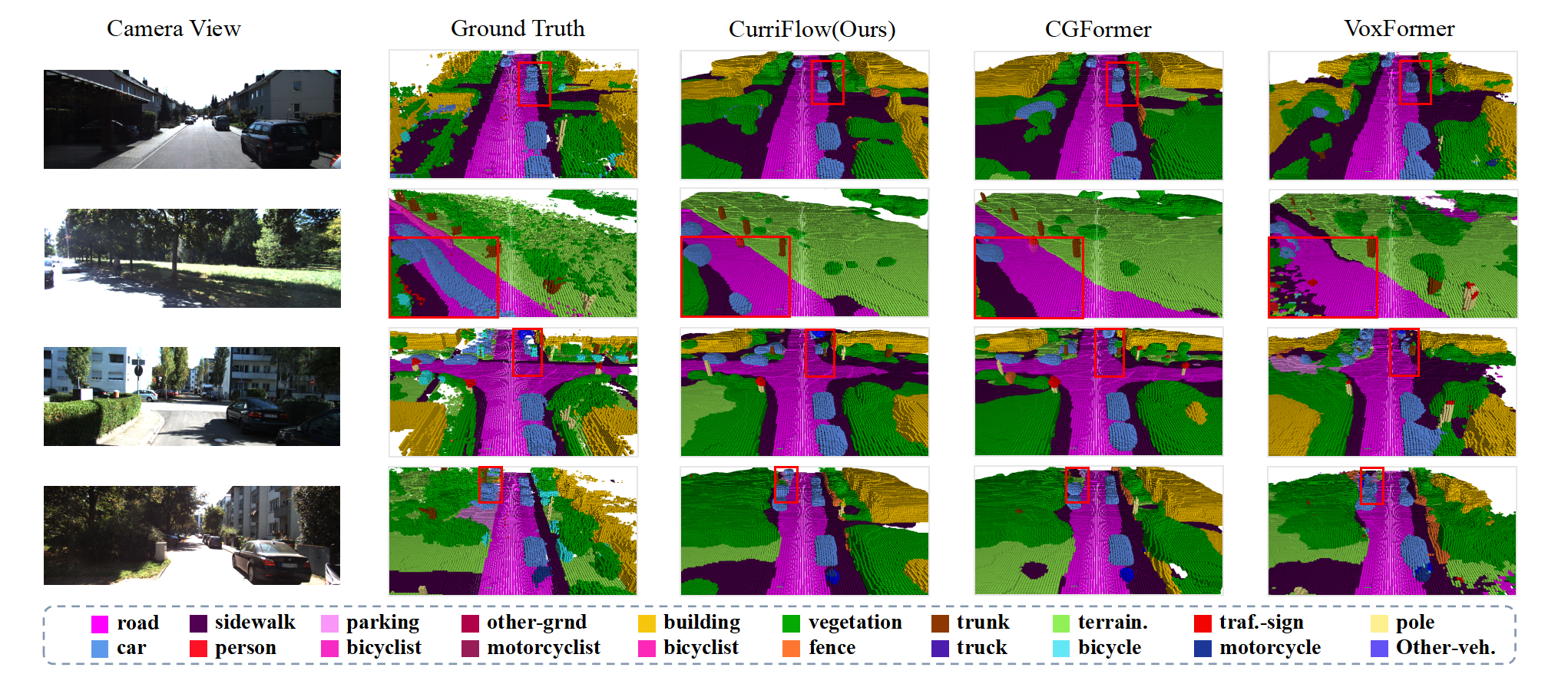}
    \caption{Qualitative comparison of scene segmentation results. The first column shows the camera view, and the subsequent columns display the voxels outputs of Ground Truth, CurriFlow (Ours), CGFormer, and VoxFormer, respectively. Red boxes highlight key differences in segmentation accuracy across the methods.}
    \label{fig:vis}
\end{figure*}

Figure~\ref{fig:vis} shows the qualitative visualization results on the SemanticKITTI test set, comparing CGFormer, VoxFormer, and CurriFlow. CurriFlow outperforms the others, especially in distant and occluded regions, due to its optical flow-guided temporal alignment, which enhances object boundaries and occlusion handling. Other methods struggle with blurred or missing voxels in occlusions. CurriFlow maintains geometric integrity and semantic clarity in dynamic scenes.
Additionally, we present the mIoU of each method at different ranges in the Table \ref{table:range}, where CurriFlow achieves the best performance across all three ranges.
\subsection{Ablation Study}
Table \ref{table:Ablation} provides a component-wise analysis of CurriFlow. The baseline model is our model without key components.
\begin{table}[!h]
\centering
\scriptsize
\setlength{\tabcolsep}{4pt}
\renewcommand{\arraystretch}{0.9}

\resizebox{\linewidth}{!}{
\begin{tabular}{c|ccccc|c}
\hline
\multirow{2}{*}{Method} & \multicolumn{2}{c|}{$\text{OFA}^2\text{Net}$}   & \multicolumn{2}{c|}{CDFNet}    & \multirow{2}{*}{G-SAM} & \multirow{2}{*}{mIoU} \\ \cline{2-5}
                        & MG & \multicolumn{1}{c|}{NCA} & CDF & \multicolumn{1}{c|}{CGA} &                        &                       \\ \hline
baseline                     &    &                          &     &                          &                        & 15.87                 \\
(1)                     & \checkmark  &                          &     &                          &                        & 16.28                 \\
(2)                     &    & \checkmark                       &     &                          &                        & 16.12                 \\
(3)                     & \checkmark  & \checkmark                        &     &                          &                        & 16.45                 \\
(4)                     & \checkmark  & \checkmark                       & \checkmark   &                          &                        & 16.66                 \\
(5)                     & \checkmark  & \checkmark                       & \checkmark   & \checkmark                        &                        & 16.74                 \\
(6)                     & \checkmark  & \checkmark                       & \checkmark   & \checkmark                        & \checkmark                      & 16.89      \\\hline          
\end{tabular}}
\caption{Ablation study of components on SemanticKITTI validation set. MG: mask gate blocks. NCA: neighborhood cross attention. DF: depth fusion. CGA3D: confidence gated attention 3D. G-SAM: Grounded-SAM. FTE: FFTOccEncoder.}
\label{table:Ablation}

\end{table}
\paragraph{Ablation on Optical Flow Alignment with Attention Network.}
Directly stacking frames without higher-level processing negatively impacts performance, due to varying camera viewpoints and redundant temporal information. After applying Mask optical flow warping and Gate , features show improved representational power, achieving an mIoU of 16.28. Using NCA alone, without the Gate Blocks, resulted in a noticeable decrease in mIoU. This highlights that relying solely on NCA without explicitly filtering out unreliable regions leads to less accurate segmentation. Therefore, the inclusion of the mask gate, which selects reliable regions, plays a crucial role in improving the performance by focusing on more trustworthy areas.
\begin{table}[!h]
\centering
\scriptsize
\setlength{\tabcolsep}{10pt}
\renewcommand{\arraystretch}{0.7}

\resizebox{\linewidth}{!}{
\begin{tabular}{cccc|c}
\hline
\multicolumn{4}{c|}{\textbf{Input}} & \multirow{2}{*}{mIoU} \\
t-1   & t-2  & t-3  & t-4  &                       \\ \hline
\checkmark     &      &      &      & 16.26                 \\
\checkmark     & \checkmark    &      &      & 16.89                 \\
\checkmark     & \checkmark    & \checkmark    &      & 16.62                 \\
\checkmark     & \checkmark   & \checkmark    & \checkmark    & 16.51                 \\ \hline         
\end{tabular}}
\caption{Comparison of mIoU with different numbers of input frames.}
\label{tab:temp}
\label{table:Ablation1}

\end{table}
\begin{table}[!h]
\centering
\scriptsize
\setlength{\tabcolsep}{10pt}
\renewcommand{\arraystretch}{0.7}

\resizebox{\linewidth}{!}{
\begin{tabular}{cccc}
\hline
\multicolumn{1}{c|}{\multirow{2}{*}{\textbf{Method}}} & \multicolumn{3}{c}{\textbf{mIoU}}                   \\
\multicolumn{1}{c|}{}                                 & \textbf{12.8 m} & \textbf{25.6 m} & \textbf{51.2 m} \\ \hline
MonoScene                                             & 12.3            & 12.2            & 11.3            \\
VoxFormer-T                                           & 21.6            & 18.4            & 13.4            \\
HASSC-T                                               & 24.10           & 20.27           & 14.74           \\
H2GFormer-T                                           & 23.43           & 20.37           & 14.29           \\
SGN-T                                                 & 25.70           & 22.02           & 15.32           \\
CurriFlow                                             & \textbf{25.9}   & \textbf{22.4}   & \textbf{16.89}  \\ \hline

\end{tabular}}
\caption{Comparison of mIoU at different ranges for various methods.}
\label{table:range}

\end{table}
\paragraph{Ablation on Curriculum-Guided Depth Fusion Network.}
he curriculum-guided fusion enables the model to leverage accurate LiDAR supervision in early training while gradually adapting to noisy stereo inputs, improving convergence stability and generalization. Compared to fixed-weight fusion or stereo-only training, this strategy yields better 3D reconstruction, especially in occluded or uncertain regions. The proposed Curriculum-guided Depth Fusion (CDF) module improves mIoU by 0.21, and the addition of CGA3D brings a further 0.08 gain, demonstrating effective depth volume enhancement. It is important to note that using CGA3D alone without CDF is meaningless due to the absence of depth information. Therefore, this combination is not included in the table.

\paragraph{Ablation on Grounded-SAM}
Grounded-SAM provides an improvement of 0.15 in mIoU, highlighting its effectiveness in enhancing semantic understanding. As a pre-trained segmentation model, Grounded-SAM contributes valuable semantic priors, allowing the model to focus on relevant features. These priors guide the model in distinguishing between important foreground objects and background, thereby improving the overall performance in complex or ambiguous regions.
\paragraph{Ablation on Temporal Input}
We present the results for different input frame counts in the Table \ref{tab:temp} and observe that as the number of input frames increases, the mIoU initially improves, reaching its peak before gradually decreasing with further increases in the frame count. This phenomenon can be explained by the fact that a moderate number of input frames provides more temporal and contextual information, helping the model better capture dynamic changes and fine-grained details in the scene, thereby improving the model's accuracy. However, as the number of frames increases, the model may be influenced by redundant information, especially when high frame counts introduce irrelevant data or noise that interferes with the model’s learning process, leading to a decline in performance. Additionally, the biases in the optical flow model accumulate as the number of frames increases, amplifying errors and further affecting the model's performance.
\section{Conclusion}
We propose CurriFlow, a semantic occupancy prediction framework that leverages optical flow for temporal alignment, curriculum-guided depth fusion, and semantic distillation from pre-trained vision models. By mitigating viewpoint inconsistency, noisy depth, and occlusions, CurriFlow enhances temporal modeling, geometric robustness, and semantic understanding, achieving state-of-the-art performance on the SemanticKITTI dataset in complex dynamic scenes.

\vfill
\bibliographystyle{IEEEtran}
\bibliography{ref}

\begin{thebibliography}{10}
\providecommand{\url}[1]{#1}
\csname url@samestyle\endcsname
\providecommand{\newblock}{\relax}
\providecommand{\bibinfo}[2]{#2}
\providecommand{\BIBentrySTDinterwordspacing}{\spaceskip=0pt\relax}
\providecommand{\BIBentryALTinterwordstretchfactor}{4}
\providecommand{\BIBentryALTinterwordspacing}{\spaceskip=\fontdimen2\font plus
\BIBentryALTinterwordstretchfactor\fontdimen3\font minus \fontdimen4\font\relax}
\providecommand{\BIBforeignlanguage}[2]{{%
\expandafter\ifx\csname l@#1\endcsname\relax
\typeout{** WARNING: IEEEtran.bst: No hyphenation pattern has been}%
\typeout{** loaded for the language `#1'. Using the pattern for}%
\typeout{** the default language instead.}%
\else
\language=\csname l@#1\endcsname
\fi
#2}}
\providecommand{\BIBdecl}{\relax}
\BIBdecl

\bibitem{tian2023occ3d}
X.~Tian, T.~Jiang, L.~Yun, Y.~Mao, H.~Yang, Y.~Wang, Y.~Wang, and H.~Zhao, ``Occ3d: A large-scale 3d occupancy prediction benchmark for autonomous driving,'' \emph{Advances in Neural Information Processing Systems}, vol.~36, pp. 64\,318--64\,330, 2023.

\bibitem{li2023voxformer}
Y.~Li, Z.~Yu, C.~Choy, C.~Xiao, J.~M. Alvarez, S.~Fidler, C.~Feng, and A.~Anandkumar, ``Voxformer: Sparse voxel transformer for camera-based 3d semantic scene completion,'' in \emph{Proceedings of the IEEE/CVF conference on computer vision and pattern recognition}, 2023, pp. 9087--9098.

\bibitem{cao2022monoscene}
A.-Q. Cao and R.~De~Charette, ``Monoscene: Monocular 3d semantic scene completion,'' in \emph{Proceedings of the IEEE/CVF Conference on Computer Vision and Pattern Recognition}, 2022, pp. 3991--4001.

\bibitem{han2025multimodal}
X.~Han, S.~Chen, Z.~Fu, Z.~Feng, L.~Fan, D.~An, C.~Wang, L.~Guo, W.~Meng, X.~Zhang \emph{et~al.}, ``Multimodal fusion and vision-language models: A survey for robot vision,'' \emph{arXiv preprint arXiv:2504.02477}, 2025.

\bibitem{yu2024context}
Z.~Yu, R.~Zhang, J.~Ying, J.~Yu, X.~Hu, L.~Luo, S.-Y. Cao, and H.-L. Shen, ``Context and geometry aware voxel transformer for semantic scene completion,'' \emph{Advances in Neural Information Processing Systems}, vol.~37, pp. 1531--1555, 2024.

\bibitem{wang2024not}
S.~Wang, J.~Yu, W.~Li, W.~Liu, X.~Liu, J.~Chen, and J.~Zhu, ``Not all voxels are equal: Hardness-aware semantic scene completion with self-distillation,'' in \emph{Proceedings of the IEEE/CVF Conference on Computer Vision and Pattern Recognition}, 2024, pp. 14\,792--14\,801.

\bibitem{li2024bevformer}
Z.~Li, W.~Wang, H.~Li, E.~Xie, C.~Sima, T.~Lu, Q.~Yu, and J.~Dai, ``Bevformer: learning bird's-eye-view representation from lidar-camera via spatiotemporal transformers,'' \emph{IEEE Transactions on Pattern Analysis and Machine Intelligence}, 2024.

\bibitem{xu2022gmflow}
H.~Xu, J.~Zhang, J.~Cai, H.~Rezatofighi, and D.~Tao, ``Gmflow: Learning optical flow via global matching,'' in \emph{Proceedings of the IEEE/CVF conference on computer vision and pattern recognition}, 2022, pp. 8121--8130.

\bibitem{cho2024flowtrack}
S.~Cho, J.~Huang, S.~Kim, and J.-Y. Lee, ``Flowtrack: Revisiting optical flow for long-range dense tracking,'' in \emph{Proceedings of the IEEE/CVF Conference on Computer Vision and Pattern Recognition}, 2024, pp. 19\,268--19\,277.

\bibitem{chen2025sage}
S.~Chen, C.~Wang, R.~Xu, X.~Pei, Y.~Song, J.~Lin, W.~Xu, J.~Zhang, L.~Guo, and S.~Xu, ``Sage: Spatial-visual adaptive graph exploration for visual place recognition,'' \emph{arXiv preprint arXiv:2509.25723}, 2025.

\bibitem{song2017semantic}
S.~Song, F.~Yu, A.~Zeng, A.~Chang, M.~Savva, and T.~Funkhouser, ``Semantic scene completion from a single depth image,'' in \emph{CVPR}, 2017.

\bibitem{behley2019semantickitti}
J.~Behley, M.~Garbade, A.~Milioto, J.~Quenzel, S.~Behnke, C.~Stachniss, and J.~Gall, ``Semantickitti: A dataset for semantic scene understanding of lidar sequences,'' \emph{ICCV}, 2019.

\bibitem{caesar2020nuscenes}
H.~Caesar, V.~Bankiti, A.~H. Lang, S.~Vora, V.~E. Liong, Q.~Xu, A.~Krishnan, Y.~Pan, G.~Baldan, and O.~Beijbom, ``nuscenes: A multimodal dataset for autonomous driving,'' in \emph{Proceedings of the IEEE/CVF conference on computer vision and pattern recognition}, 2020, pp. 11\,621--11\,631.

\bibitem{huang2023tri}
Y.~Huang, W.~Zheng, Y.~Zhang, J.~Zhou, and J.~Lu, ``Tri-perspective view for vision-based 3d semantic occupancy prediction,'' in \emph{Proceedings of the IEEE/CVF conference on computer vision and pattern recognition}, 2023, pp. 9223--9232.

\bibitem{miao2023occdepth}
R.~Miao, W.~Liu, M.~Chen, Z.~Gong, W.~Xu, C.~Hu, and S.~Zhou, ``Occdepth: A depth-aware method for 3d semantic scene completion,'' \emph{arXiv preprint arXiv:2302.13540}, 2023.

\bibitem{bruhn2005lucas}
A.~Bruhn, J.~Weickert, and C.~Schn{\"o}rr, ``Lucas/kanade meets horn/schunck: Combining local and global optic flow methods,'' \emph{International journal of computer vision}, vol.~61, no.~3, pp. 211--231, 2005.

\bibitem{baker2004lucas}
S.~Baker and I.~Matthews, ``Lucas-kanade 20 years on: A unifying framework,'' \emph{International journal of computer vision}, vol.~56, no.~3, pp. 221--255, 2004.

\bibitem{dosovitskiy2015flownet}
A.~Dosovitskiy, P.~Fischer, E.~Ilg, P.~Hausser, C.~Hazirbas, V.~Golkov, P.~Van Der~Smagt, D.~Cremers, and T.~Brox, ``Flownet: Learning optical flow with convolutional networks,'' in \emph{Proceedings of the IEEE international conference on computer vision}, 2015, pp. 2758--2766.

\bibitem{ilg2017flownet}
E.~Ilg, N.~Mayer, T.~Saikia, M.~Keuper, A.~Dosovitskiy, and T.~Brox, ``Flownet 2.0: Evolution of optical flow estimation with deep networks,'' in \emph{Proceedings of the IEEE conference on computer vision and pattern recognition}, 2017, pp. 2462--2470.

\bibitem{sun2018pwc}
D.~Sun, X.~Yang, M.-Y. Liu, and J.~Kautz, ``Pwc-net: Cnns for optical flow using pyramid, warping, and cost volume,'' in \emph{Proceedings of the IEEE conference on computer vision and pattern recognition}, 2018, pp. 8934--8943.

\bibitem{teed2020raft}
Z.~Teed and J.~Deng, ``Raft: Recurrent all-pairs field transforms for optical flow,'' in \emph{European conference on computer vision}.\hskip 1em plus 0.5em minus 0.4em\relax Springer, 2020, pp. 402--419.

\bibitem{jiang2021learning}
Q.~Jiang, Z.~Xing, Z.~Chen, Z.~Zhu, and G.~Huang, ``Learning optical flow via global matching,'' in \emph{Proceedings of the IEEE/CVF Conference on Computer Vision and Pattern Recognition}, 2021, pp. 13\,886--13\,895.

\bibitem{huang2022flowformer}
Z.~Huang, X.~Shi, C.~Zhang, Q.~Wang, K.~C. Cheung, H.~Qin, J.~Dai, and H.~Li, ``Flowformer: A transformer architecture for optical flow,'' in \emph{European conference on computer vision}.\hskip 1em plus 0.5em minus 0.4em\relax Springer, 2022, pp. 668--685.

\bibitem{zhu2017flow}
X.~Zhu, Y.~Wang, J.~Dai, L.~Yuan, and Y.~Wei, ``Flow-guided feature aggregation for video object detection,'' in \emph{Proceedings of the IEEE international conference on computer vision}, 2017, pp. 408--417.

\bibitem{hassani2023neighborhood}
A.~Hassani, S.~Walton, J.~Li, S.~Li, and H.~Shi, ``Neighborhood attention transformer,'' in \emph{Proceedings of the IEEE/CVF conference on computer vision and pattern recognition}, 2023, pp. 6185--6194.

\bibitem{zhang2023completionformer}
Y.~Zhang, X.~Guo, M.~Poggi, Z.~Zhu, G.~Huang, and S.~Mattoccia, ``Completionformer: Depth completion with convolutions and vision transformers,'' in \emph{Proceedings of the IEEE/CVF conference on computer vision and pattern recognition}, 2023, pp. 18\,527--18\,536.

\bibitem{Shamsafar_2022_WACV}
F.~Shamsafar, S.~Woerz, R.~Rahim, and A.~Zell, ``Mobilestereonet: Towards lightweight deep networks for stereo matching,'' in \emph{Proceedings of the IEEE/CVF Winter Conference on Applications of Computer Vision (WACV)}, January 2022, pp. 2417--2426.

\bibitem{hu2018squeeze}
J.~Hu, L.~Shen, and G.~Sun, ``Squeeze-and-excitation networks,'' in \emph{Proceedings of the IEEE conference on computer vision and pattern recognition}, 2018, pp. 7132--7141.

\bibitem{philion2020lift}
J.~Philion and S.~Fidler, ``Lift, splat, shoot: Encoding images from arbitrary camera rigs by implicitly unprojecting to 3d,'' in \emph{European conference on computer vision}.\hskip 1em plus 0.5em minus 0.4em\relax Springer, 2020, pp. 194--210.

\bibitem{xia2022vision}
Z.~Xia, X.~Pan, S.~Song, L.~E. Li, and G.~Huang, ``Vision transformer with deformable attention,'' in \emph{Proceedings of the IEEE/CVF conference on computer vision and pattern recognition}, 2022, pp. 4794--4803.

\bibitem{ren2024grounded}
T.~Ren, S.~Liu, A.~Zeng, J.~Lin, K.~Li, H.~Cao, J.~Chen, X.~Huang, Y.~Chen, F.~Yan \emph{et~al.}, ``Grounded sam: Assembling open-world models for diverse visual tasks,'' \emph{arXiv preprint arXiv:2401.14159}, 2024.

\bibitem{lmscnet}
\BIBentryALTinterwordspacing
L.~Roldão, R.~de~Charette, and A.~Verroust-Blondet, ``Lmscnet: Lightweight multiscale 3d semantic completion,'' 2020. [Online]. Available: \url{https://arxiv.org/abs/2008.10559}
\BIBentrySTDinterwordspacing

\bibitem{wei2016convolutional}
S.-E. Wei, V.~Ramakrishna, T.~Kanade, and Y.~Sheikh, ``Convolutional pose machines,'' in \emph{Proceedings of the IEEE conference on Computer Vision and Pattern Recognition}, 2016, pp. 4724--4732.

\bibitem{liang2022bevfusion}
T.~Liang, H.~Xie, K.~Yu, Z.~Xia, Z.~Lin, Y.~Wang, T.~Tang, B.~Wang, and Z.~Tang, ``Bevfusion: A simple and robust lidar-camera fusion framework,'' \emph{Advances in Neural Information Processing Systems}, vol.~35, pp. 10\,421--10\,434, 2022.

\bibitem{pan2023uniocc}
M.~Pan, L.~Liu, J.~Liu, P.~Huang, L.~Wang, S.~Zhang, S.~Xu, Z.~Lai, and K.~Yang, ``Uniocc: Unifying vision-centric 3d occupancy prediction with geometric and semantic rendering,'' \emph{arXiv preprint arXiv:2306.09117}, 2023.

\bibitem{sinha2017surfnet}
A.~Sinha, A.~Unmesh, Q.~Huang, and K.~Ramani, ``Surfnet: Generating 3d shape surfaces using deep residual networks,'' in \emph{Proceedings of the IEEE conference on computer vision and pattern recognition}, 2017, pp. 6040--6049.

\bibitem{kirillov2023segment}
A.~Kirillov, E.~Mintun, N.~Ravi, H.~Mao, C.~Rolland, L.~Gustafson, T.~Xiao, S.~Whitehead, A.~C. Berg, W.-Y. Lo \emph{et~al.}, ``Segment anything,'' in \emph{Proceedings of the IEEE/CVF international conference on computer vision}, 2023, pp. 4015--4026.

\bibitem{caron2021emerging}
M.~Caron, H.~Touvron, I.~Misra, H.~J{\'e}gou, J.~Mairal, P.~Bojanowski, and A.~Joulin, ``Emerging properties in self-supervised vision transformers,'' in \emph{Proceedings of the IEEE/CVF international conference on computer vision}, 2021, pp. 9650--9660.

\bibitem{radford2021learning}
A.~Radford, J.~W. Kim, C.~Hallacy, A.~Ramesh, G.~Goh, S.~Agarwal, G.~Sastry, A.~Askell, P.~Mishkin, J.~Clark \emph{et~al.}, ``Learning transferable visual models from natural language supervision,'' in \emph{International conference on machine learning}.\hskip 1em plus 0.5em minus 0.4em\relax PmLR, 2021, pp. 8748--8763.

\bibitem{tan2019efficientnet}
M.~Tan and Q.~Le, ``Efficientnet: Rethinking model scaling for convolutional neural networks,'' in \emph{International conference on machine learning}.\hskip 1em plus 0.5em minus 0.4em\relax PMLR, 2019, pp. 6105--6114.

\bibitem{morimitsu2025dpflow}
H.~Morimitsu, X.~Zhu, R.~M. Cesar, X.~Ji, and X.-C. Yin, ``Dpflow: Adaptive optical flow estimation with a dual-pyramid framework,'' in \emph{Proceedings of the Computer Vision and Pattern Recognition Conference}, 2025, pp. 17\,810--17\,820.

\bibitem{ke2023segment}
L.~Ke, M.~Ye, M.~Danelljan, Y.-W. Tai, C.-K. Tang, F.~Yu \emph{et~al.}, ``Segment anything in high quality,'' \emph{Advances in Neural Information Processing Systems}, vol.~36, pp. 29\,914--29\,934, 2023.

\bibitem{duhamel1990fast}
P.~Duhamel and M.~Vetterli, ``Fast fourier transforms: a tutorial review and a state of the art,'' \emph{Signal processing}, vol.~19, no.~4, pp. 259--299, 1990.

\bibitem{liu2021swin}
Z.~Liu, Y.~Lin, Y.~Cao, H.~Hu, Y.~Wei, Z.~Zhang, S.~Lin, and B.~Guo, ``Swin transformer: Hierarchical vision transformer using shifted windows,'' in \emph{Proceedings of the IEEE/CVF international conference on computer vision}, 2021, pp. 10\,012--10\,022.

\end{thebibliography}
\nocite{*}
\end{document}